\documentclass[10pt,leqno]{article}
\usepackage[a4paper]{geometry}

\usepackage{biblatex}
\usepackage{graphicx}
\usepackage[caption=false]{subfig}
\usepackage{float}
\usepackage{stfloats}
\usepackage{amsmath}
\usepackage{enumitem}
\setdescription{nosep} 
\setlist{noitemsep} 

\usepackage[algo2e, ruled, noend]{algorithm2e}
\SetKwArray{prev}{prev}
\SetKwArray{curr}{curr}
\SetKwFunction{swap}{swap}
\SetKwFunction{co}{co}
\SetKwFunction{li}{li}
\SetKwFunction{kwmin}{min}
\SetKwData{nextstart}{next\_start}
\SetKwData{prevpruningpoint}{prev\_pruning\_point}
\SetKwData{pruningpoint}{pruning\_point}
\SetKwData{ub}{ub}
\SetKwData{lb}{lb}

\usepackage{hyperref}

\DeclareMathOperator{\DTW}{DTW}
\DeclareMathOperator{\cost}{cost}

\begin{document}
\title{\Large Early Abandoning PrunedDTW and its application to similarity search}
\author{Matthieu Herrmann\thanks{Monash University, Melbourne, Australia.} \and Geoffrey I. Webb\footnotemark[1]}
\date{%
Faculty of Information Technology, Monash University, Melbourne, Australia.
}
\maketitle

\begin{abstract} \small\baselineskip=9pt
  The Dynamic Time Warping (``DTW'') distance is  widely used  in time series analysis,
  be it for classification, clustering or similarity search.
  However, its quadratic time complexity prevents it from scaling.
  Strategies, based on early abandoning DTW or skipping its computation altogether thanks to lower bounds,
  have been developed for certain use cases such as nearest neighbour search.
  But vectorization and approximation aside,
  no advance was made on DTW itself until recently with the introduction of PrunedDTW.
  This algorithm, able to prune unpromising alignments, was later fitted with early abandoning.
  We present a new version of PrunedDTW, ``EAPrunedDTW'',
  designed with early abandon in mind from the start, and able to early abandon faster than before.
  We show that EAPrunedDTW significantly improves the computation time of similarity search in the UCR Suite,
  and renders lower bounds dispensable.
\end{abstract}

\section{\label{sec-introduction}Introduction}
DTW \cite{sakoe_dynamic_1978} is one of the most important similarity measure in time series analysis,
extensively used in clustering~\cite{luczak_hierarchical_2016},
classification~\cite{wang_experimental_2013}, and similarity search~\cite{rakthanmanon_searching_2012}.
Classification with one nearest neighbour under DTW (``NN1-DTW'') was a widely used benchmark for a long time,
and while state of the art classifiers are now significantly more accurate, some of them actually embed a NN1-DTW
classifier as a component~\cite{lines_time_2015,lines_time_2018,lucas_proximity_2019,shifaz_ts-chief_2020}.

A naive DTW implementation has space and time complexities both undesirably quadratic.
How to take care of the former is well known~\cite{mueen_extracting_2016}
(and we do so in this article, cf~section~\ref{sec-background}),
while taking care of the later is more challenging.
The common ``early abandoning'' strategy first tries to avoid DTW computations,
and then tries to early abandon them.
But it can only speed up scenarios where an upper bound is available (e.g.~NN1~search), and not DTW itself.
DTW has been sped up through vectorization~\cite{xiao_parallelizing_2013,mueen_extracting_2016}
or approximation~\cite{salvador_toward_2007}%
\footnote{Using constraint bands or approximation generally leads to different results than
with ``exact DTW'', and thus cannot be considered as being faster versions of DTW.},
and more recently by PrunedDTW~\cite{silva_speeding_2016},
an algorithm developed for scenarios where early abandoning isn't applicable
(e.g.~when all pairwise distances are required).
It was later applied to similarity search in the UCR USP Suite~\cite{silva_speeding_2018},
fitted with \emph{some} early abandon capabilities used in conjunction with lower bounding.
However, the core of the algorithm was not revisited.

In this article, we present ``EAPrunedDTW'',
our version of PrunedDTW specifically designed to efficiently and effectively early abandon.
We incorporated our algorithm in the UCR Suite and reproduced the experiments from~\cite{silva_speeding_2018},
showing a significant speed improvement over previous implementations based on DTW and PrunedDTW.
Moreover, we show that using lower bounds, even if useful, becomes dispensable.
This is especially relevant for other distances sharing the same structure as DTW
(see~\cite{marteau_time_2009}) for which no efficient and tight lower bounds exist.

The next section introduces DTW and related work.
We then present our algorithm in two parts, sections~\ref{sec-presentation-1} and~\ref{sec-presentation-2}.
Experiments follow in section~\ref{sec-experimentations}, and we conclude in section~\ref{sec-conclusion}.

\section{\label{sec-background}Background and related work}
Given two times series $S$ and $T$ of respective length $l_S$ and $l_T$,
and a $\cost$ function (usually the squared Euclidean distance)
over data points themselves denoted by application (e.g.~$S(i)$),
the distance $\DTW^{S,T}$ between $S$ and $T$ is defined by:
\begin{figure}[H]
\setlength{\abovedisplayskip}{0pt}
\setlength{\belowdisplayskip}{0pt}
\begin{alignat}{2}
 & \hspace{-3em} \DTW^{S,T}       && = \DTW^{S,T}(l_S, l_T) \\
 & \hspace{-3em} \DTW^{S,T}(0,0)  && = 0 \label{eq:DTW:border1}\\
 & \hspace{-3em} \DTW^{S,T}(i,0)  && = +\infty \label{eq:DTW:border2}\\
 & \hspace{-3em} \DTW^{S,T}(0,j)  && = +\infty \label{eq:DTW:border3}\\
 & \hspace{-3em} \DTW^{S,T}(i,j)  && =
    \cost(S(i), T(j)) + \min\left\{
      \begin{array}{l}
        \DTW^{S,T}(i-1, j)   \\
        \DTW^{S,T}(i, j-1)   \\
        \DTW^{S,T}(i-1, j-1)
      \end{array}
    \right.\label{eq:DTW:main}
\end{alignat}
\caption{\label{eq:DTW}Equations defining $\DTW^{S,T}$}
\end{figure}

\begin{figure}[t]
  \centering
  \subfloat[\label{fig:DTW:Matrix}
  DTW matrix with warping path for $S$ and $T$.
  We have $\DTW^{S,T}=M_{\DTW}^{S,T}(6,6)=9$.
  ]{
    \includegraphics[trim=0 35 0 0, clip,width=0.45\textwidth]{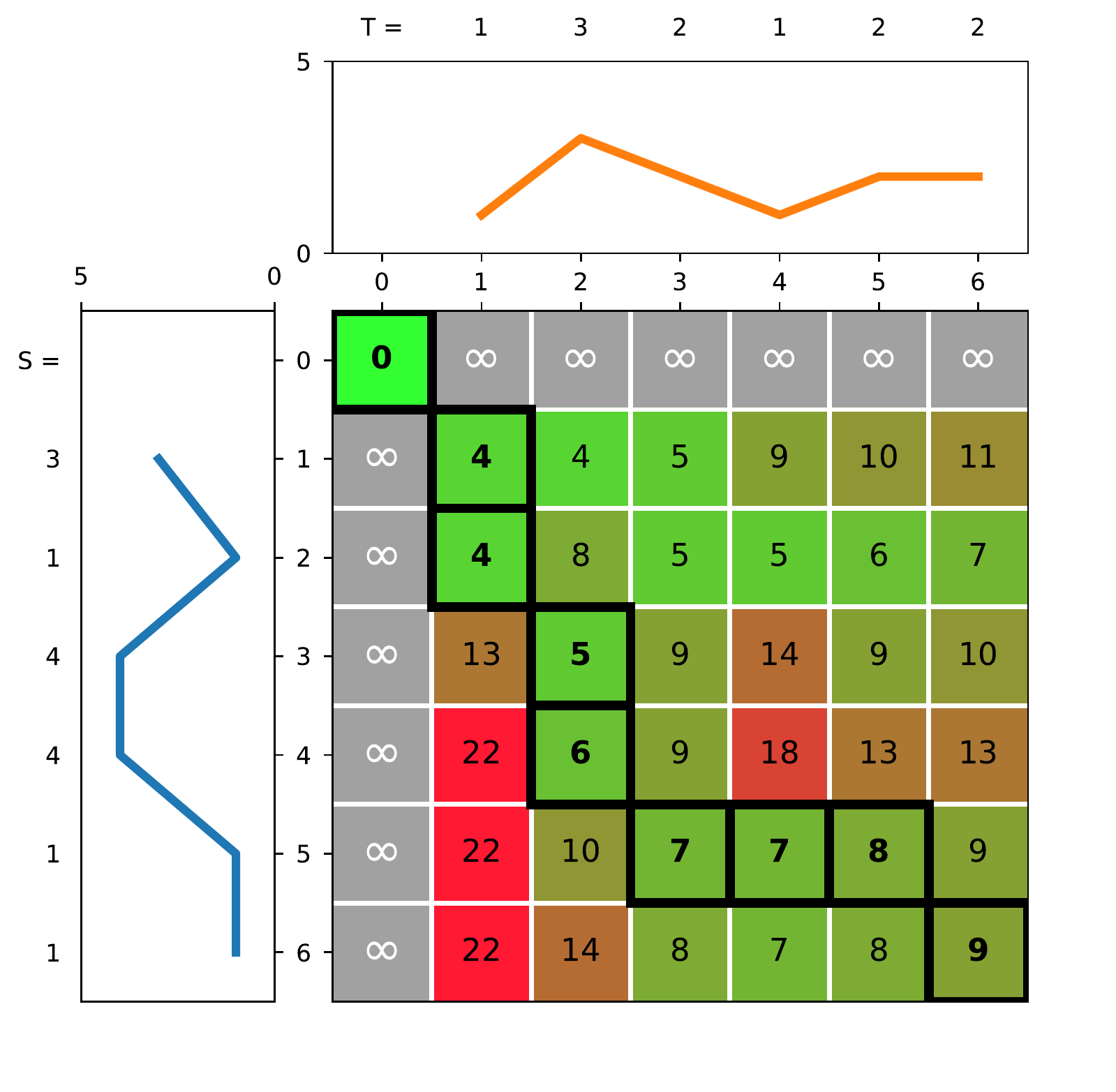}
  }%
  \hspace{\fill}
  \subfloat[\label{fig:DTW:Alignments}
  DTW alignements for $S$ and $T$ with their costs.
  Their sum is $\DTW^{S,T}=9$.]{%
    \includegraphics[width=0.45\textwidth]{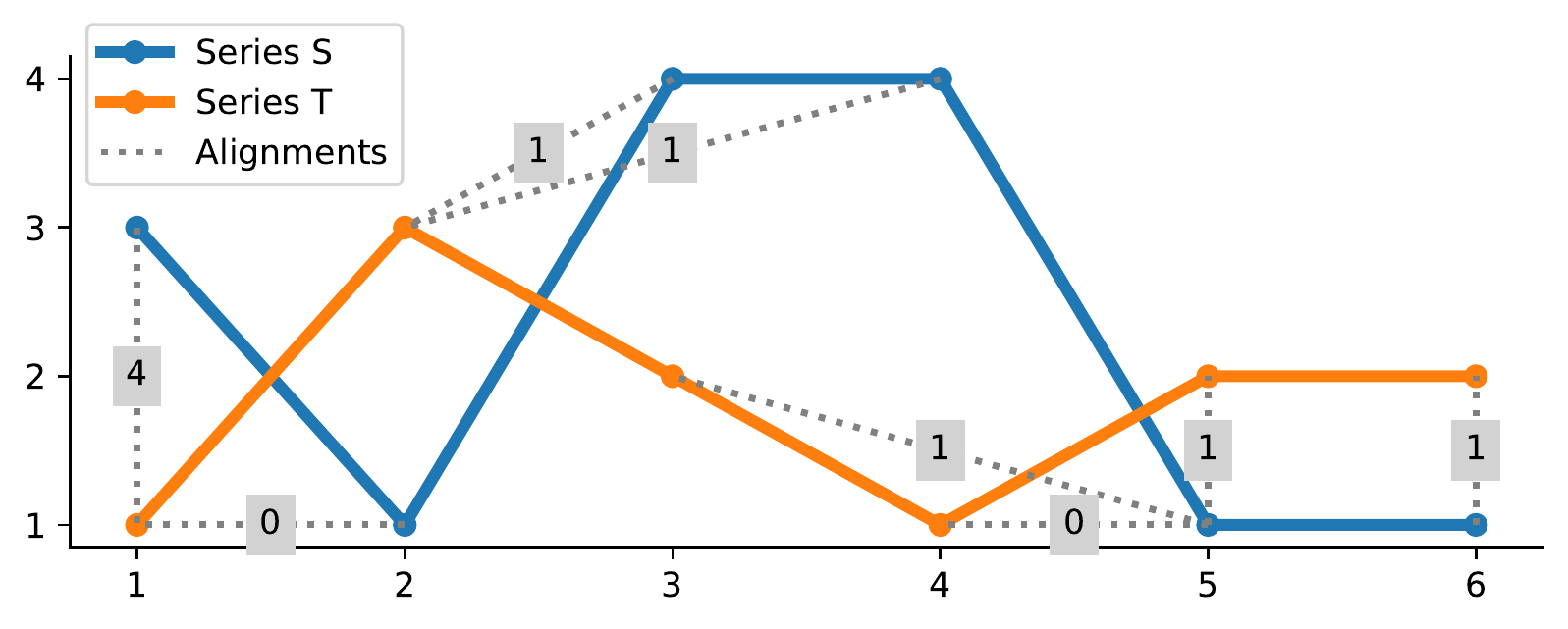}
  }
  \caption{\label{fig:DTW}
    DTW matrix $M_{\DTW}^{S,T}$ with warping path and alignments for \mbox{$S = (3, 1, 4, 4, 1, 1)$} and \mbox{$T = (1, 3, 2, 1, 2, 2)$}.
    A warping path's cell $(i,j)$ means that $S(i)$ is aligned with $T(j)$.
  }
\end{figure}

We define by \mbox{$M_{\cost}^{S,T}(i,j) = \cost(S(i), S(j))$} the cost matrix of $S$ and $T$.
A warping path in $M_{\cost}^{S,T}$ is a continuous and monotonous path connecting $(1,1)$ to $(l_S, l_T)$.
The value $\DTW^{S,T}$ is the cost of the optimal (cheapest)
warping path\footnote{There may be several optimal paths with the same cost.},
which also represents the cost of the optimal non linear alignment between $S$ and $T$.
Such an alignment is illustrated by Figure~\ref{fig:DTW:Alignments}
for \mbox{$S = (3, 1, 4, 4, 1, 1)$} and \mbox{$T = (1, 3, 2, 1, 2, 2)$}.
The constraints on the path lead to the extremities (at $(1,1)$ and $l_S, l_T)$ being aligned
and no alignments crossing each other (monotonicity).

Let $S(1\dots{i})$ be the prefix of $S$ of length~$i$.
We defined the DTW matrix of the series $S$ and $T$ by
\mbox{$M_{\DTW}^{S,T}(i,j) = \DTW^{S(1\dots{i}),T(1\dots{j})}$}.
Figure~\ref{fig:DTW:Matrix} shows the matrix and optimal warping path for $S$ and $T$
(this and subsequent figures are colored from intense green for the minimum value 0,
through to intense red for the maximum value 22).
We are only interested in the DTW matrix's last cell at $(l_S, l_T)$,
i.e.~the value of $\DTW^{S,T}$.
This cell can be computed without keeping the full matrix in memory.
Equation~\ref{eq:DTW:main} tells us that a cell's value
only depends on its ``top'', ``top left diagonal'', and ``left'' neighbours.
In a line by line, left to right scan, the two firsts are obtained from the previous line,
while the later is the previously computed cell.
Hence, we only need to keep the current and previous line in memory,
limiting the space complexity to $O(l)$ as described by Algorithm~\ref{alg:DTW}.

Commenting Algorithm~\ref{alg:DTW}.
Memory allocation is limited by matching the line dimension with the shortest series.
Both arrays, representing the previous and current lines of the matrix,
have an extra cell for the vertical border ($j=0$) and are initialized to $\infty$.
The horizontal border ($i=0$)  is initially represented by the $\curr$ array,
which is swapped for the $\prev$ array prior to any computation (line~\ref{alg:dtw:swap}).
This is why we initialize the border cell $(0,0)$
in the extra cell of the $\curr$ array (line~\ref{alg:dtw:init0}).
This cell has to be set back to $\infty$ after the first iteration.
We repeatedly do so (line~\ref{alg:dtw:inf}),
which is avoidable by computing the first line before the loop.
However, our goal is to introduce a structure we will build upon:
a form of this assignment is required later.

\begin{algorithm2e}[t]
    \small
    \SetAlgoLined
    \LinesNumbered
    \DontPrintSemicolon
    \SetKwData{c}{c}
    \SetKwFunction{cost}{cost}
    \KwIn{the time series $S$ and $T$}
    \KwResult{$\DTW^{S,T}$}
    \co $\leftarrow$ shortest series between $S$ and $T$\;
    \li $\leftarrow$ longest series between $S$ and $T$\;
    (\prev, \curr) $\leftarrow$ arrays of length $l_{\co}+1$ filled with $\infty$\;
    $\curr{0} \leftarrow 0 $\; \label{alg:dtw:init0}
    \For{$i \leftarrow 1$ \KwTo $l_{\li}$}{
      \swap{\prev, \curr}\; \label{alg:dtw:swap}
      \curr{0} $\leftarrow \infty$\; \label{alg:dtw:inf}
      \For{$j \leftarrow 1$ \KwTo $l_{\co}$}{
        $\c \leftarrow \cost{\li{i}, \co{j}} $\;
        $\curr{j} \leftarrow \c + \kwmin{\curr{j-1}, \prev{j}, \prev{j-1}}$\;
      }
    }
    \Return{\curr{$l_{\co}$}}\;
    \caption{\label{alg:DTW}$O(n)$ space complexity $\DTW^{S,T}$.}
\end{algorithm2e}

\subsection{The Warping Window}
DTW is commonly parametrized by a warping window $w$ constraining
the warping path to deviate by at most $w$ cells from the diagonal~\cite{sakoe_dynamic_1978}.
A window of 0 is equivalent to the $\cost$ function being applied to the successive elements of the two series
(e.g. the squared Euclidean distance), while a window of $\min(l_{S},l_T)$ is equivalent to DTW.
The experiments section~\ref{sec-experimentations} require a warping window,
but for clarity's sake we present our algorithms for DTW only.

\subsection{Early Abandoning}
Early abandoning is commonly used to speed up scenarios requiring nearest neighbour search,
like NN1 classification or similarity search.
In those scenarios, we test a query $S$ against several candidates $T_1, T_2, \dots T_m$.
After computing $v_1 = \DTW^{S,T_1}$, we have a first upper bound $\ub=v_1$
on the DTW cost between $S$ and its actual nearest neighbour,
i.e. the nearest neighbour will have a cost lower than or equal to $\ub$.
We tighten $\ub$ every time we find a better candidate.
It follows that we can early abandon the computation of $\DTW^{S,T_k}$
as soon as we can determine that $v_k$ will \emph{strictly} exceed $\ub$.
This strictness condition prevents from early abandoning ties.
To implement DTW with early abandon
it is enough to get the the minimum value of each $\curr$ line,
and stop if it is strictly greater then $\ub$.
The UCR Suite also exploits the Keogh lower bound to further tighten $\ub$ \cite{rakthanmanon_searching_2012}.

Another form of early abandoning is lower bounding:
The idea is to cheaply compute a lower bound $\lb(S, T_k)\leq\DTW^{S,T_k}$
and to skip any further computation if $\lb(S, T_k)>\ub$.
Several lower bounds have been developed, and developing them is still an active field of research.
The most commonly used lower bounds are LB\_Kim and LB\_Keogh~\cite{mueen_extracting_2016}.

\subsection{PrunedDTW}
Early abandoning requires an upper bound on the best candidates.
PrunedDTW~\cite{silva_speeding_2016} was designed for scenarios where no such upper bound exists,
e.g. when all pairwise distances are required.
In those cases, we can only speed-up DTW itself.
PrunedDTW determines that some cells from the DTW matrix cannot be part of the optimal warping path
and skips their computation.
To do so, it relies on a pruning threshold which is an upper bound on the current alignment's cost
(and not on the cost of the alignment with the best candidates).
This threshold can only depend on the series being aligned.
PrunedDTW uses the squared Euclidean distance, which corresponds to the diagonal in the DTW matrix\footnote{The cost of any valid alignment is an upper bound.
Either the alignment is the optimal one, or the optimal one will have a lower cost.},
which may be quite loose.

The next logical step was to use PrunedDTW in scenarios with upper bounds
on best candidates as they are usually tighter.
The UCR Suite has been modified to use PrunedDTW, giving birth to the UCR USP Suite \cite{silva_speeding_2018}.
However, PrunedDTW inherits features from its origins as a pruning algorithm, and fails to fully exploit the possibilities of early abandoning. In the current work we seek to exploit those opportunities to the full.

\subsection{Overheads and Run-Time Speed}
The computation of DTW relies of two nested ``tight'' loops:
the inner body is small but repeated many times.
Early abandoning and pruning techniques modify those loops with extra checking and bookkeeping,
adding a bit of ``overhead'' per iteration.
The expectation is that the time saved by those techniques outweigh their overhead.
It follows that for the same gain, the algorithm with the smallest overhead will be the most efficient.
By carefully decomposing our algorithm in several stages, we seek to minimize these overheads while maximising pruning.

Comparing the run-time speed of different algorithms seems simple:
run the candidates several times under the same conditions and compare the fastest (i.e. less noisy) results.
There is not much to discuss if the differences are substantial (regardless of the direction),
and clearly due to algorithmic changes.
However, we are not comparing algorithms, but implementations of those algorithms.
A ``fast'' implementation of a ``slow'' algorithm may be the best performer until we reach large enough data.
To combat this problem we embed our DTW algorithm in the UCR Suite and make minimal modifications to the code.

\section{\label{sec-presentation-1}Pruning and Early Abandoning from the left}
\begin{figure}[!t]
  \centering
  \subfloat[][\label{fig:DTW:EAMatrix:UB9}DTW Matrix for $S$ and $T$ with $\ub=9$.
  The arrows represent the dependencies of the cell $(4,1)$.]{%
    \includegraphics[trim=0 35 0 0, clip, width=0.45\textwidth]{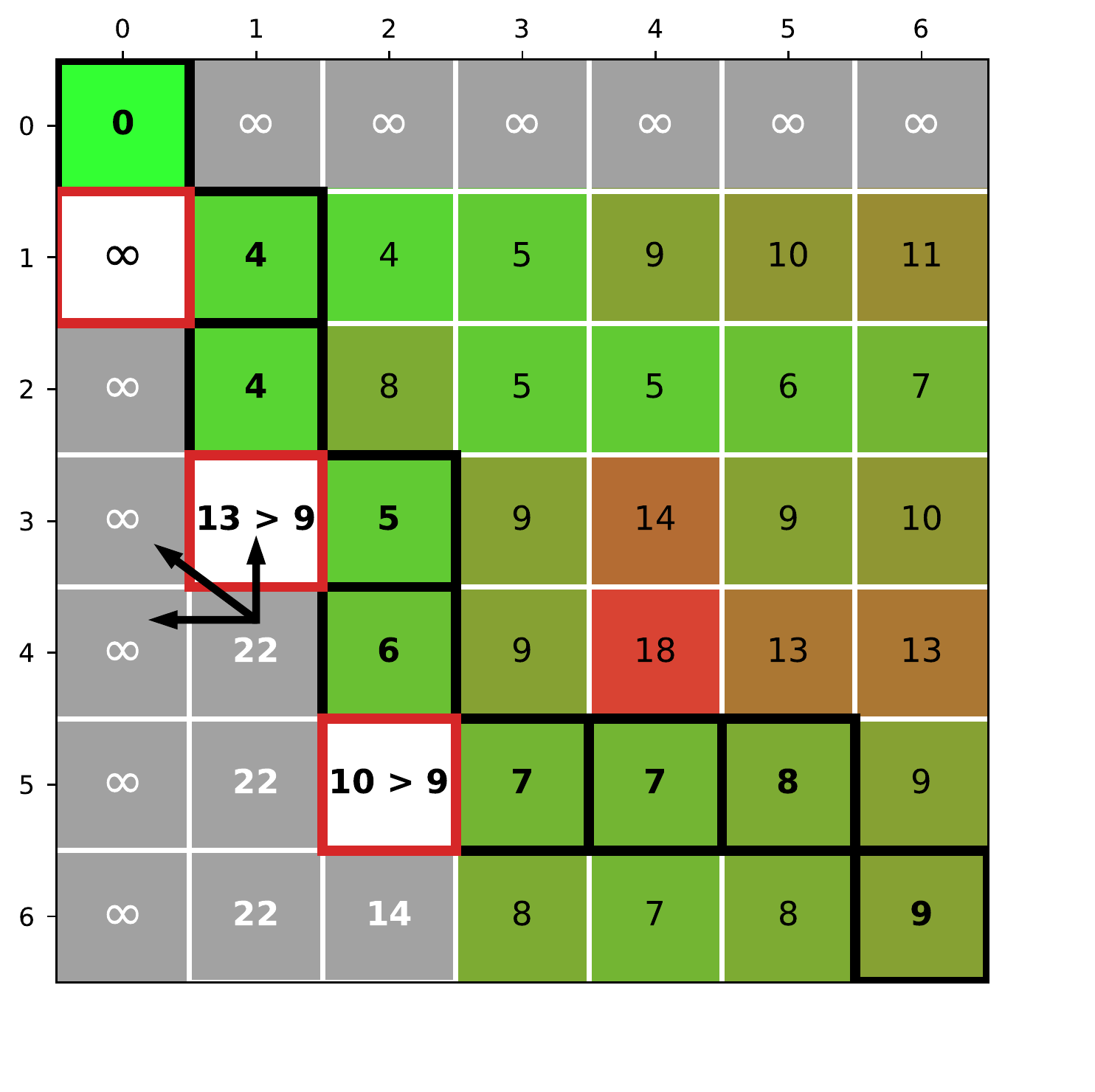}
  }%
  \hspace{\fill}
  \subfloat[][\label{fig:DTW:EAMatrix:UB6}DTW Matrix for $S$ and $T$ with $\ub=6$.
    Early abandon occurs at the end of the fifth line.]{%
    \includegraphics[trim=0 35 0 0, clip, width=0.45\textwidth]{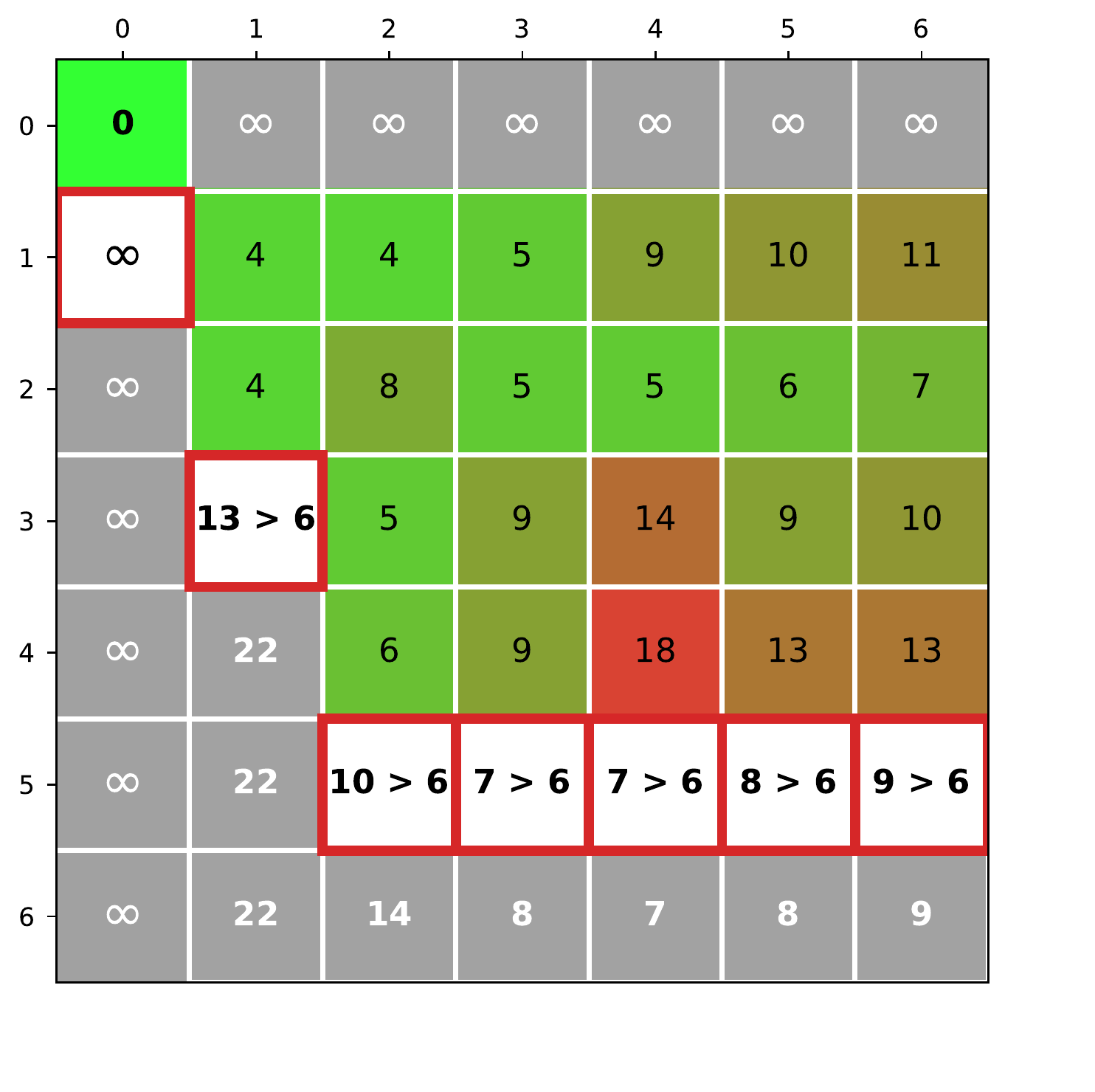}
  }
  \caption{\label{fig:DTW:EAMatrix}Representation of $M_{\DTW}^{S,T}$ ``from the left'' pruning for
  \mbox{$S = (3, 1, 4, 4, 1, 1)$} and \mbox{$T = (1, 3, 2, 1, 2, 2)$} with two different upper bounds.
  }
\end{figure}

In the previous section, we show how to compute $\DTW^{S,T}$ line by line.
Now, we show how to prune the computation ``from the left''.
As we scan a line, we look at discarding as many as possible of the leftmost cells of the DTW matrix.
We do so by extending the ``left border'' up to the leftmost cell of the current line that could
potentially be part of an optimal warping path.
The Figures~\ref{fig:DTW:EAMatrix:UB9} and \ref{fig:DTW:EAMatrix:UB6} present
the pruning and early abandoning scenarios for the same $S$ and $T$ as before,
using upper bounds of $\ub=9$ and $\ub=6$.
As $\DTW^{S,T}=9$, no early abandoning occurs in the first case, but it does in the second.

\begin{algorithm2e}[!t]
    \small
    \SetAlgoLined
    \LinesNumbered
    \DontPrintSemicolon
    \SetKwData{c}{c}
    \SetKwFunction{cost}{cost}
    \KwIn{the time series $S$ and $T$, an upper bound $\ub$}
    \KwResult{$\DTW^{S,T}$}
    \co $\leftarrow$ shortest series between $S$ and $T$\;
    \li $\leftarrow$ longest series between $S$ and $T$\;
    (\prev, \curr) $\leftarrow$ arrays of length $l_{\co}+1$ filled with $\infty$\;
    $\curr{0} \leftarrow 0 $\;
    $\nextstart \leftarrow 1$\; \label{alg:DTWEA:init_ns}
    \For{$i \leftarrow 1$ \KwTo $l_{\li}$}{
      \swap{\prev, \curr}\;
      $j \leftarrow \nextstart$\;
      $\curr{j-1} \leftarrow \infty$\; \label{alg:DTW:inf}
      \While{$j=\nextstart \wedge j \leq l_{\co}$}{ \label{alg:DTWEA:cond1}
        $\c \leftarrow \cost{\li{i}, \co{j}} $\;
        $\curr{j} \leftarrow \c + \kwmin{\prev{j}, \prev{j-1}}$\; \label{alg:DTWEA:min}
        \lIf{$\curr{j}>\ub$}{$\nextstart \leftarrow \nextstart+1$}
        $j\leftarrow j+1$\;
      }
      \lIf{$j>l_{co}$}{\Return{$\infty$}} \label{alg:DTWEA:ea}
      \While{$j \leq l_{\co}$}{
        $\c \leftarrow \cost{\li{i}, \co{j}} $\;
        $\curr{j} \leftarrow \c + \kwmin{\curr{j-1}, \prev{j}, \prev{j-1}}$\;
        $j\leftarrow j+1$\;
      }
    }
    \Return{\curr{$l_{\co}$}}\;
    \caption{\label{alg:DTWEA}Pruning from the left.}
\end{algorithm2e}

Looking at Figure~\ref{fig:DTW:EAMatrix}, the grey cells are never computed (border or pruned cells),
and the white cells with red edges represent ``discard points''.
Given a line and an upper bound $\ub$,
discard points represent a continuous sequence of cells greater than $\ub$, starting at the left border.
By monotonicity, all columns below discard points are guaranteed to be greater than $\ub$.
It follows that those columns can be skipped for the rest of the computation,
and we do so by starting the next line after the current line's last discard point%
\footnote{This explain why the Algorithm~\ref{alg:DTWEA} uses the ``\nextstart''
variable rather than a ``discard point'' one}.
Ignoring the top border for now (we will take care of it in the next section),
you can  easily check that all the dependencies of the grey cells are either another grey cell, or a discard point.
Note that $(1,0)$ is a discard point while $(2,0)$ is pruned.
In this regard, the first discard point and pruned column naturally match the left border.

Looking at the discard point at $(3,1)$ in Figure~\ref{fig:DTW:EAMatrix:UB9} with $\ub=9$,
we see that all the dependencies (black arrows) of $(4,1)$ are greater than $\ub$, and the same goes until $(6,1)$.
It follows that we can start the following lines on the second columns, at $(3,2)$,
effectively skipping the cells below $(3,1)$ and intuitively extending the border by one cell to the right.
Note that the continuity condition, that all cells to the left of a discard point must also exceed $\ub$, is paramount.
Without it, the cell at $(3,4)$ with value $14>9$ would be marked as a discard point
and wrongfully prevent the optimal warping path from being computed.

Now looking at Figure~\ref{fig:DTW:EAMatrix:UB6} with $\ub=6$,
we see that the discard points reach the end of the fifth line, leading to early abandoning.
Under this strategy, early abandoning is the natural consequence of extending the border over the full length of a line.
This is a very effective way of early abandoning, requiring liitle bookkeeping.
In contrast, PrunedDTW~\cite{silva_speeding_2018} maintains the minimum across a row and abandons when that minimum exceeds $\ub$.

Algorithm~\ref{alg:DTWEA} shows how this works, in two stages.
Lines now start at index \nextstart instead of index 1.
The first line still starts at $1$, so we initialize \nextstart to $1$,
and then update it by $j+1$ after encountering a discard point at $(i, j)$.
The first stage, represented by the first inner loop, computes successive discard points on the same line.
The condition on line~\ref{alg:DTWEA:cond1} enforces our continuity requirement:
if \nextstart did not increase, the previous cell is not a discard point, and so neither is the current cell.
Also, while in this loop, we know that a cell's left neighbour is greater then \ub
and thus can be left out when looking for the smallest dependency (see line~\ref{alg:DTWEA:min}).
In that regard, line~\ref{alg:DTW:inf} seems unnecessary --
it isn't as it also represents an assignment to the top left diagonal at the next iteration.
We then have to check how we exited the first inner loop.
If we reached the end of the line, we early abandon (line~\ref{alg:DTWEA:ea}),
else we go into the second stage of the algorithm, embodied by the second inner loop, which is a normal DTW computation.

\section{\label{sec-presentation-2}Early Abandoned PrunedDTW}
In the previous section, we intuitively extended the left border to the right,
leading to early abandoning when it reaches the end of the line.
Unfortunately, we cannot do the same thing for the top border.
This is because our strategy implicitly relies on the computation order (line by line)
by moving forward where we start a line (updating $\nextstart$).
In a line by line scan ``moving where a column starts'' does not make sense.
However, we can move backward where we end a line pruning cells up to its actual end.
As before, we ensure that all pruned cells would be greater than an upper bound $\ub$
by ensuring that their dependencies are greater than $\ub$.
Adding this ``pruning from the right'' to the algorithm~\ref{alg:DTWEA} gives us the full EAPrunedDTW algorithm.

\begin{figure}[!t]
  \centering
  \subfloat[][\label{fig:DTW:EAPrunedMatrix:UB9}DTW Matrix for $S$ and $T$ with $\ub=9$]{%
    \includegraphics[trim=0 35 0 0, clip, width=0.45\textwidth]{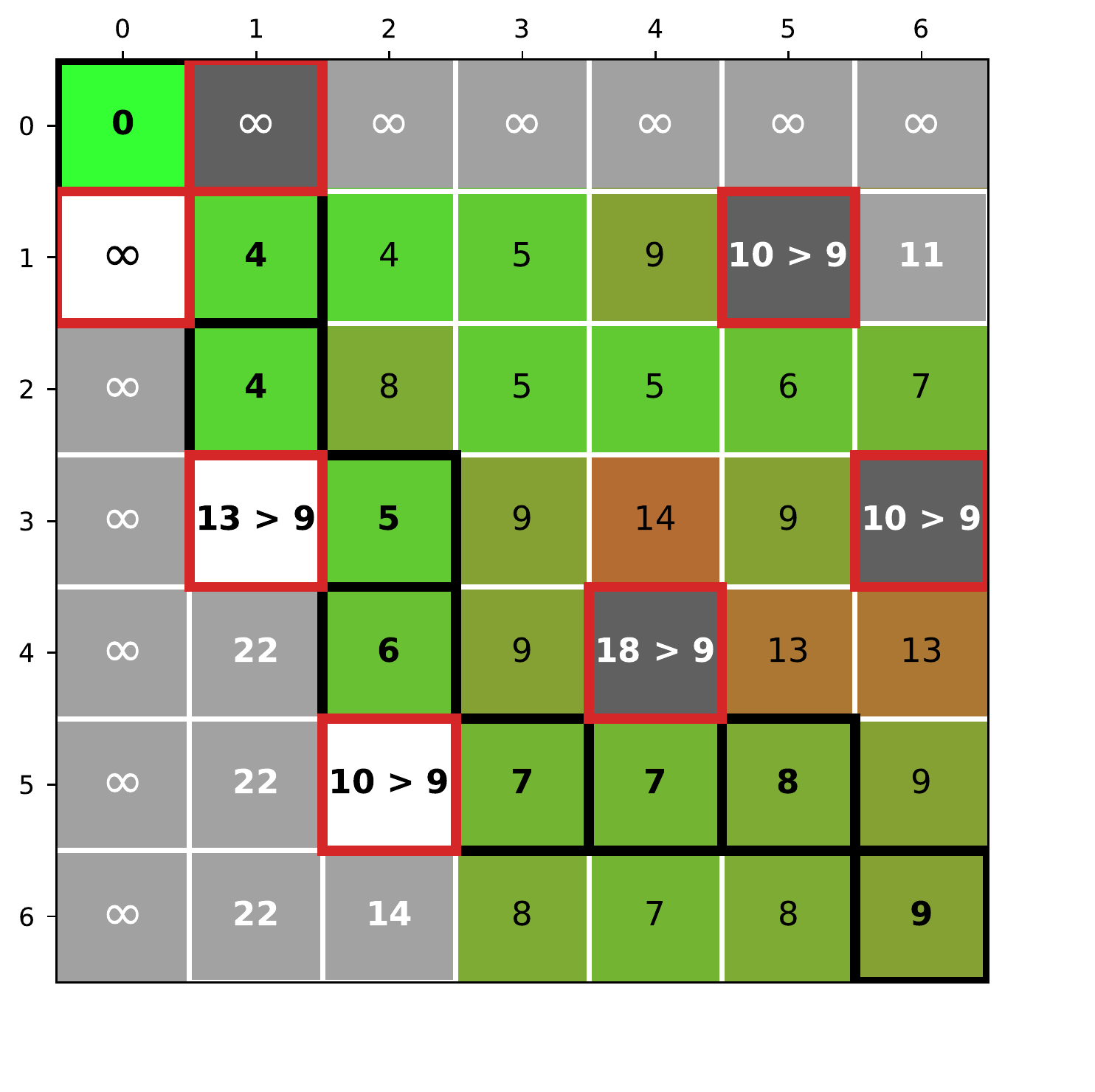}
  }%
  \hspace{\fill}
  \subfloat[][\label{fig:DTW:EAPrunedMatrix:UB6}DTW Matrix for $S$ and $T$ with $\ub=6$.
  The arrows show the dependencies actually evaluated]{%
    \includegraphics[trim=0 35 0 0, clip, width=0.45\textwidth]{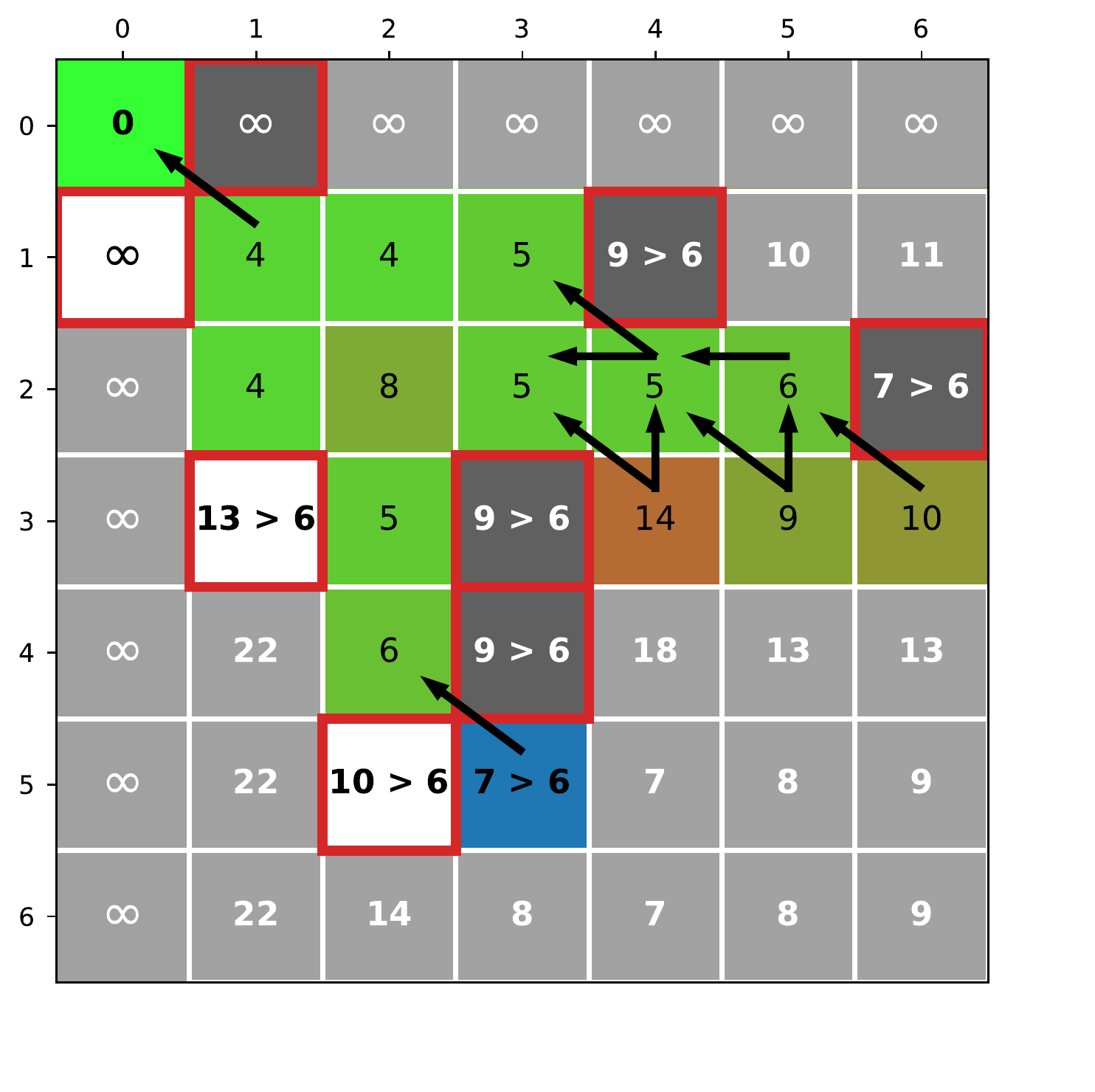}
  }
  \caption{\label{fig:DTWEAPruned}Illustration of EAPrunedDTW.
    The blue cell represents the point of early abandoning, white cells represent discard points,
    and dark gray cells represent pruning points.
  }
\end{figure}

Figures~\ref{fig:DTW:EAPrunedMatrix:UB9} and~\ref{fig:DTW:EAPrunedMatrix:UB6}
present the EAPrunedDTW scenarios for $S$ and $T$, with $\ub=9$ and $\ub=6$.
The blue cell shows an early abandoning position,
and the black cells with red border are ``pruning points''.
In a line, a pruning point starts a continuous sequence of cells greater than $\ub$
until the actual end of the line.
In this regard, we have another natural match,
now between the first pruning point at $(0,1)$ and the top border,
as for all cells after $(0, 1)$ we indeed have $\infty>\ub$.

Examining Figure~\ref{fig:DTW:EAPrunedMatrix:UB6} with $\ub=6$,
the cell $(2,2)=8>\ub$ is not a pruning point as it is followed by cells below $\ub$.
Notice that we can prune after the pruning points at $(1, 4)$ and $(4,3)$, but not after $(3,3)$.
This is because a pruning point at $(i, j)$ carries information for the next line:
it tells us that a cell at $(i+1, j'>j)$ can only depend on its left neighbour,
its top and top left neighbours being greater than $\ub$.
It follows that all cells following a cell $(i+1, j')>\ub$ will be greater than $\ub$.
This allow us to both create a pruning point at $(i+1, j')$ and to prune the rest of the line.
This explains why we prune after $(4,3)$ but not after $(3,3)$: the former can be pruned thanks to the latter!
Moreover, cells after $(3,3)$ must be computed in order to check the continuity requirement,
allowing $(3,3)$ to be a pruning point.
Notice how, contrary to the discard points that extend the left border once and for all,
pruning points create a ``right border'' that can go back and forth.

Given a pruning point at $(i, j)$, we discussed what happens to the cells at $(i+1, j'>j)$.
We now discuss the case of the cell immediately below, at $(i+1, j)$.
This point is of particular interest as it is where our left and right borders can ``collide''.
Either it depends on both its left and top left neighbours (e.g.~$(2,4)$),
or only on its top left neighbour if it follows a discard point (e.g.~$(5,3)$ and $(1,1)$).
In the latter case, we early abandon if the cell is greater than $\ub$ (e.g. the blue cell at $(5,3)$)
as this would put $\nextstart$ in a pruned area, creating a ``border collision''.
Note that PrunedDTW does prune ``from the left'' and ``from the right'',
but does not take the border collision into account, early abandoning by checking the minimum value of the line.
This allows EAPrunedDTW to abandon earlier than PrunedDTW.

\begin{algorithm2e}[!t]
    \small
    \SetAlgoLined
    \LinesNumbered
    \DontPrintSemicolon
    \SetKwData{c}{c}
    \SetKwFunction{cost}{cost}
    \KwIn{the time series $S$ and $T$, an upper bound $\ub$}
    \KwResult{$\DTW^{S,T}$}
    \co $\leftarrow$ shortest series between $S$ and $T$\;
    \li $\leftarrow$ longest series between $S$ and $T$\;
    (\prev, \curr) $\leftarrow$ arrays of length $l_{\co}+1$ filled with $\infty$\;
    $\curr{0} \leftarrow 0 $\;
    $\nextstart \leftarrow 1$\;
    $\prevpruningpoint \leftarrow 1$\; \label{alg:EAPrunedDTW:initpp}
    $\pruningpoint \leftarrow 0$\;
    \For{$i \leftarrow 1$ \KwTo $l_{\li}$}{
      \swap{\prev, \curr}\;
      $j \leftarrow \nextstart$\;
      $\curr{j-1} \leftarrow \infty$\;
      \While{$j=\nextstart \wedge j < \prevpruningpoint$}{
        $\c \leftarrow \cost{\li{i}, \co{j}} $\;
        $\curr{j} \leftarrow \c + \kwmin{\prev{j}, \prev{j-1}}$\; \label{alg:EAPrunedDTW:2m}
        \lIf{$\curr{j}\leq\ub$}{$\pruningpoint \leftarrow j+1$}\lElse{$\nextstart \leftarrow \nextstart+1$}
        $j\leftarrow j+1$\;
      }
      \While{$j < \prevpruningpoint$}{
        $\c \leftarrow \cost{\li{i}, \co{j}} $\;
        $\curr{j} \leftarrow \c + \kwmin{\curr{j-1}, \prev{j}, \prev{j-1}}$\;
        \lIf{$\curr{j}\leq\ub$}{$\pruningpoint \leftarrow j+1$}
        $j\leftarrow j+1$\;
      }
      \If(\tcp*[h]{At \prevpruningpoint}){$j\leq{}l_{\co}$}{ \label{alg:EAPrunedDTW:pp}
        $\c \leftarrow \cost{\li{i}, \co{j}} $\;
        \If{$j=\nextstart$}{ \label{alg:EAPrunedDTW:testtp}
          $\curr{j} \leftarrow \c + \prev{j-1}$\;
          \lIf{$\curr{j}\leq\ub$}{$\pruningpoint \leftarrow j+1$}\lElse{\Return{$\infty$}}
        }
        \Else{
          $\curr{j} \leftarrow \c + \kwmin{\curr{j-1}, \prev{j-1}}$\;
          \lIf{$\curr{j}\leq\ub$}{$\pruningpoint \leftarrow j+1$}
        }
        $j \leftarrow j+1$\;
      }
      \uElseIf{$j=\nextstart$}{\Return{$\infty$}} \label{alg:EAPrunedDTW:ea}
      \While{$j = \pruningpoint \wedge j \leq l_{\co}$}{ \label{alg:EAPrunedDTW:postpp}
        $\curr{j} \leftarrow \c + \curr{j-1}$\; \label{alg:EAPrunedDTW:1m}
        \lIf{$\curr{j}\leq\ub$}{$\pruningpoint \leftarrow j+1$}
        $j \leftarrow j+1$\;
      }
      $\prevpruningpoint \leftarrow \pruningpoint$\;
    }
    \lIf{$\prevpruningpoint > l_{\co}$}{ \Return{\curr{$l_{\co}$}} }
    \lElse{\Return{$\infty$}}
    \caption{\label{alg:EAPrunedDTW}Early Abandoning PrunedDTW.}
\end{algorithm2e}

EAPrunedDTW is described in algorithm~\ref{alg:EAPrunedDTW},
in four stages, building upon algorithm~\ref{alg:DTWEA}.
We introduce two new variables.
The first one, \prevpruningpoint, represents the pruning point of the previous line and is initialized to 1
(for the first pruning point at $(0,1)$, line~\ref{alg:EAPrunedDTW:initpp}).
The second one, \pruningpoint, represents the pruning point being computed in the current line,
and is initialized to 0.
This concurs with our update strategy: we assume that a cell is a pruning point unless proven otherwise.
In other words, if the current cell is less than $\ub$, we assume that the next one is a pruning point.
Either it is, and \pruningpoint being left untouched will end up correctly set,
or it is not and \pruningpoint will be updated again.

The first two stages, embodied by the first two inner loops, are similar to the ones in Algorithm\ref{alg:DTWEA}.
However, the loops now update \pruningpoint, only go up to \prevpruningpoint, and we early abandon in the third stage.
The third stage corresponds to the column of the previous pruning point (line~\ref{alg:EAPrunedDTW:pp}).
If we are still within bounds, we check if the previous cell is a discard point
($\nextstart=j$ at line~\ref{alg:EAPrunedDTW:testtp} tests if we reached this position while advancing $\nextstart$,
i.e.~while encountering discard point),
we only look at the diagonal, early abandoning if the computed value is greater than $\ub$.
Else, we compute the DTW value based on the left and diagonal.
In both cases we update \pruningpoint.
If we are not within bound, we reached the end of the line.
If we did so with the first loop (again tested with $\nextstart=j$),
we are in the same early abandoning situation as in Algorithm~\ref{alg:DTWEA},
and we early abandon (line~\ref{alg:EAPrunedDTW:ea}).

We finally reach the fourth stage, after the column \prevpruningpoint (line~\ref{alg:EAPrunedDTW:postpp}).
The only possible dependency is now on the left,
and we can skip the rest of the line as soon as we encounter a value greater than $\ub$.
We do this by checking \pruningpoint against $j$, ensuring that it is being updated.
The final check before returning ensures that we computed the last cell of the line in the last iteration,
as we may have stopped the computation in the last stage.

Decomposing the algorithm in several stages is the main underlying contribution of this paper.
Whereas the standard DTW loop used by PrundedDTW requires that the minimum of three previous values be computed for each new cell of the matrix, our decomposition allows many cells to be updated considering only one (line~\ref{alg:EAPrunedDTW:1m}) or two (line~\ref{alg:EAPrunedDTW:2m}) previous values, saving considerable computation.

\section{\label{sec-experimentations}Experiments}

We created the ``UCR MON Suite''
\footnote{UCR from Monash University\newline\hspace*{1.8em}\url{https://github.com/MonashTS/UCR-Monash}},
available online with the experimental results and analysis script.
Just like the ``UCR USP Suite''
\footnote{UCR from the Universidade de São Paulo\newline\hspace*{1.8em}\url{https://sites.google.com/view/ucruspsuite}}
which uses PrunedDTW,
it is a modification of the original ``UCR Suite''\footnote{\url{http://www.cs.ucr.edu/~eamonn/UCRsuite.html}}.
It uses our EAPrunedDTW algorithm  with support for a warping window and upper bound tightening
(present in both the UCR and the UCR USP suites).
Given EAPrunedDTW's efficiency, it is also interesting to assess how it performs without lower bounding,
so we added the fourth variant ``UCR MON nolb''.
Note that this last variant cannot tighten the upper bound, as this relies on information computed by lower bounds.
We compare the four suites on the same computer equipped with an AMD Opteron 6338P at 2.3Ghz (RAM is not limiting).

We reproduced the experiments from the UCR USP paper~\cite{silva_speeding_2018} over the same datasets.
See~\cite{silva_speeding_2018} for a detailed description, and how they have been obtained.
The data originated from:
\begin{description}
    \item[\cite{bachlin_wearable_2010}] Freezing of Gait (FoG)
    \item[\cite{pettersen_soccer_2014}] Athletic performance monitoring (Soccer)
    \item[\cite{reiss_introducing_2012}] Physical activity monitoring (PAMAP2)
    \item[\cite{goldberger_physiobank_2000,moody_impact_2001}] MIT-BIH Arrhythmia Database (ECG)
    \item[\cite{murray_data_2015}] Electrical load measurements (REFIT)
    \item[\cite{kachuee_cuff-less_2015}] Photoplethysmography (PPG)
\end{description}

Each dataset is composed of a long reference $R$ series, and 5 queries $Q_{1 \leq n \leq 5}$ of length 1024.
For each $Q_n$, the algorithm locates the closest subsequence in $R$
under several conditions of query's length and warping window ratio of that length.
A query's length can be 128, 256, 512 and 1024 (taking a prefix for lengths smaller than 1024),
and the window's ratio is taken among $\{0.1, 0.2, 0.3, 0.4, 0.5\}$.
This generates 20 experiments per query, 100 experiments per dataset and so 600 experiments per algorithm.
We note that our results for the UCR and UCR USP suites are in line with those published in~\cite{silva_speeding_2018}.

\begin{figure*}[!p]
  \centering
  \subfloat[][\label{fig:exp:lenght}Average runtimes by varying the query length.
    Plotted values represent the average runtime over 5 queries and varying window ratio.
  ]{%
    \includegraphics[width=0.81\linewidth]{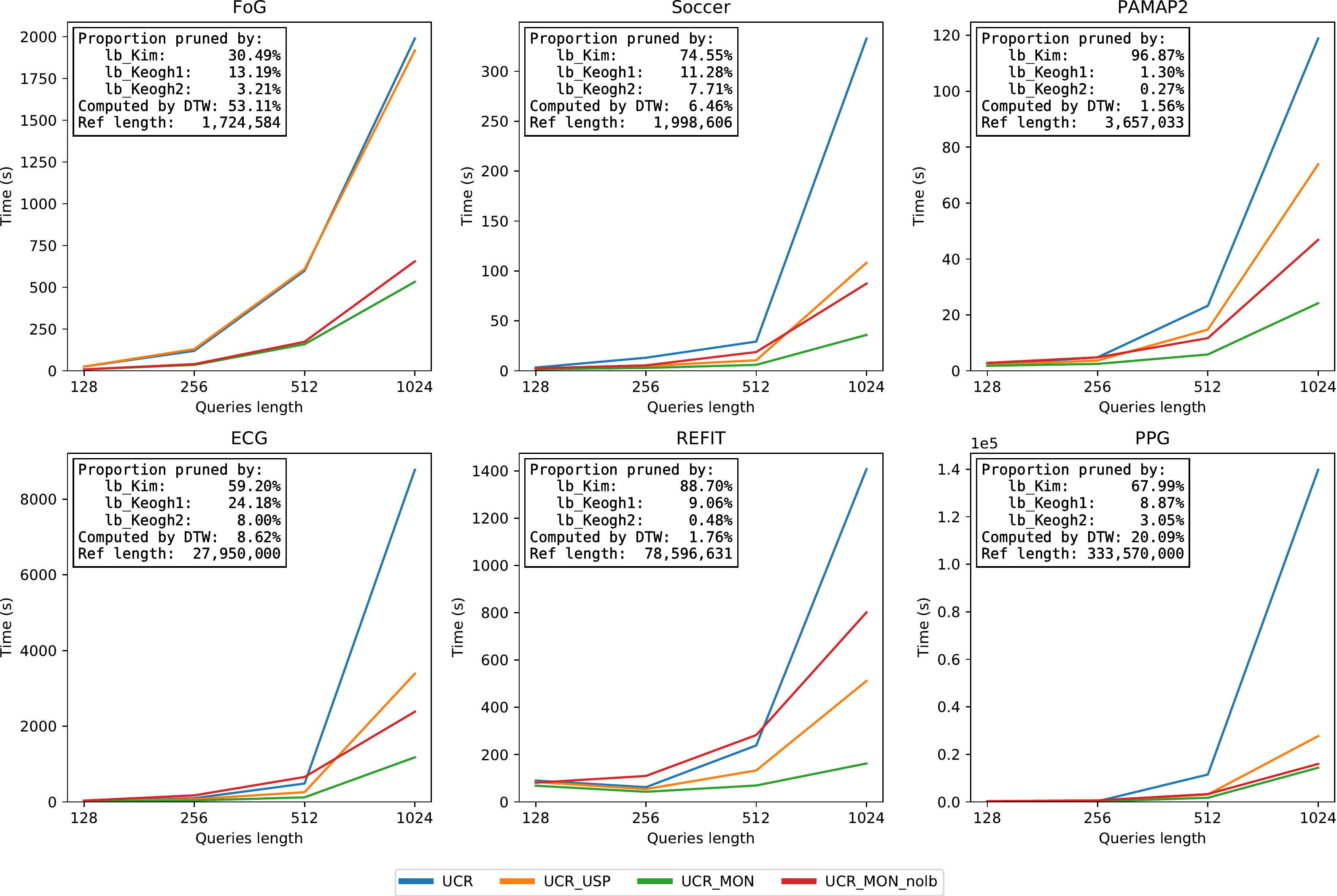}
  }%
  \vspace{1ex}
  \subfloat[][\label{fig:exp:wratio}Average runtimes by varying the window ratio.
    Plotted values represent the average runtime over 5 queries and varying query length.
  ]{%
    \includegraphics[width=0.81\linewidth]{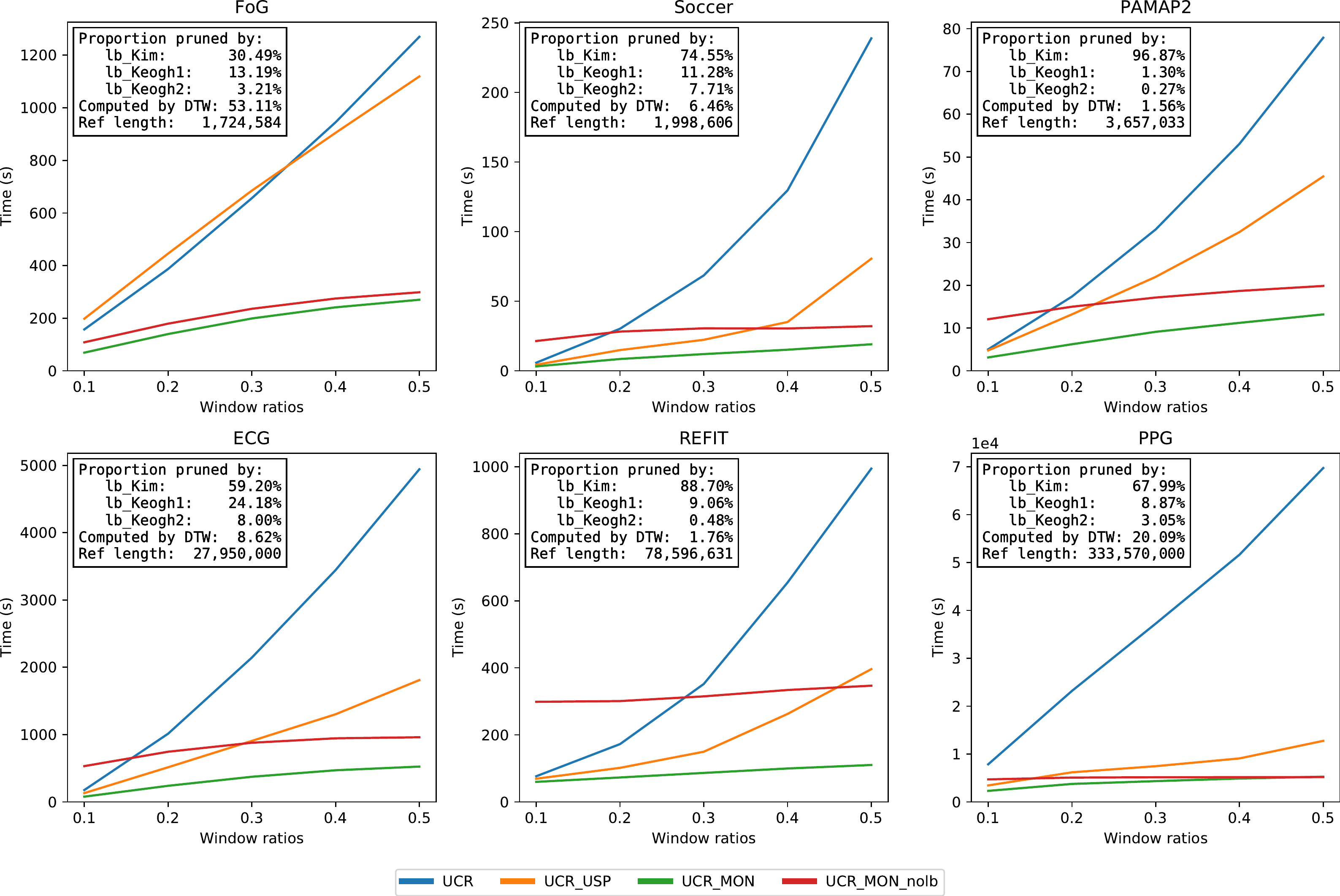}
  }
  \caption{\label{fig:exp}
    Average runtimes per dataset of UCR, UCR USP, UCR MON and UCR MON without lower bound, by query length and window ratios.
    Also show effectiveness of lower bounds pruning per datasets, applicable for all but UCR\_MON\_nolb which is 100\% DTW.
  }
\end{figure*}

We present the results in Figure~\ref{fig:exp}.
For each data set we indicate the proportion of the DTW computations
that are pruned by each of the three lower bounds used in the UCR Suite,
and the remaining proportion for which the DTW computation is performed.
Note that our algorithm is only applied to these latter cases.
The more computations that are pruned, the less opportunity our algorithm has to demonstrate its relative capacity. 

UCR MON is overall the fastest, completing the experiments in $473150$ seconds (5 days 11h25m50s), 
which is $8.778$ times faster than UCR ($4153309$ seconds, i.e.{} 48 days 1h41m49s),
and $2.036$ times faster than UCR USP
($963251$ seconds, i.e.{} 11 days 3h34m11s, $4.312$ times faster than UCR).
Its advantage is clear for long series (length 1024, Figure \ref{fig:exp:lenght}),
where it is faster than UCR from 3.73 times (FoG) up to 9.72 times (PPG),
and is faster than UCR USP from 1.93 (PPG) up to 3.6 (FoG).
In some cases the overhead is not compensated by the gain,
and our algorithm is slower but the difference stays overall small.
Our algorithm is slower than UCR in 44 ($7.3\%$) cases over 600
by an average of $0.971$ seconds, and at most by $9.06$ seconds.
It is slower than UCR USP in 66 ($11\%$) cases
by an average of $1$ seconds, and at most by $10.98$ seconds.
In comparison, UCR USP is slower than UCR in 108 ($18\%$) cases,
by an average of $55.11$ seconds, and at most by $986.15$ seconds.

The most interesting results come from UCR MON nolb.
They show that lower bounds are dispensable,
UCR MON nolb completing the experiments in $644652$ seconds (7 days 11h04m12s).
It is faster overall than both UCR (speedup $6.443$) and UCR USP (speedup $1.494$).
It is however slower than UCR in 364 ($60.06\%$) cases, by an average of $174.372$ seconds,
and at most $6392$ seconds.
Higher ratios of DTW computation lead to higher differences, especially as the window size and length increase.
A major exception is the REFIT dataset: even if it has a higher ratio of DTW computations than PAMAP2,
UCR MON nolb only starts to be competitive with a length of 1024.
Another interesting observation is that varying the window size has less impact on our algorithms than the others.
Again, this is very remarkable with REFIT.
We explain this by the following:
at first, a small window (around $0.1$, $0.2$) gives a fast result without leaving room for a lot of pruning.
As the window increases (around $0.3$, $0.4$), more and more pruning becomes possible, 
compensating both for the associated overhead and the increased numbers of cells to compute.
This continues up to a point (around $0.4$, $0.5$) where increasing the window is nearly fully compensated by pruning,
with most extra cells being pruned.
Further increases to the window result in ever more pruning, having little overall impact on runtime.
Interestingly, UCR USP does not exhibit this behaviour.

\section{\label{sec-conclusion}Conclusion and future work}

We present EAPrunedDTW, a new version of the PrunedDTW algorithm that supports highly effective pruning and early abandoning.
We implemented it in the UCR Suite,
and show it being currently the overall fastest available way of performing similarity search.
The main innovations of EAPrunedDTW are to use the collision of the left and right borders to trigger early abandoning,
saving the need to keep track of the minimum value so far,
and a decomposition that allows many cells to be updated by considering only one or two previous values
as opposed to the three previous values always required by PrunedDTW.

We also show that similarity search is tractable without using lower bounds, opening it to  distance measures other than DTW.
Indeed, many elastic measures share the same structure as DTW \cite{marteau_time_2009},
only differing in their cost function, without having lower bounds that are as efficient to compute, if any.
In essence we should be able to speed up most of elastic distances,
improving computation time in most scenarios where best candidates are required.
This includes nearest neighbour classification, which in turn is widely used in several ensemble classifiers.
We note that the Elastic Ensemble was removed from a revision of Hive Cote due to its computation time \cite{bagnall_tale_2020},
something that may not be necessary with our algorithm.

Our next steps will be to adapt our algorithm to other elastic distances and implement them in ensemble classifiers
such as Elastic Ensemble~\cite{lines_time_2015}, Proximity Forest~\cite{lucas_proximity_2019}
and TS-Chief~\cite{shifaz_ts-chief_2020}.

\printbibliography

\end{document}